\documentclass[letterpaper]{jpconf}
\usepackage{graphicx}
\usepackage{booktabs}
\usepackage{lmodern}
\usepackage{array, multirow}
\usepackage{mathtools, amsmath, braket}
\usepackage{url}
\begin{document}
\title{On-board classification of underwater images using hybrid classical-quantum CNN based method}

\author{Sreeraj Rajan Warrier$^1$, D Sri Harshavardhan Reddy$^2$, Sriya Bada$^2$,  Rohith Achampeta$^{3}$, Sebastian Uppapalli$^3$ and  Jayasri Dontabhaktuni*$^1$}

\address{$^1$Department of Physics, Mahindra University, Hyderabad, 500043, Telangana, India}
\address{$^2$Department of Computer Science, Mahindra University, Hyderabad, 500043, Telangana, India}
\address{$^3$Department of Mechanical \& Aerospace Engineering, Mahindra University, Hyderabad, 500043, Telangana, India}

\ead{jayasri.d@mahindrauniversity.edu.in}

\begin{abstract}
Underwater images taken from autonomous underwater vehicles (AUV's) often suffer from low light, high turbidity, poor contrast, motion-blur and excessive light scattering and hence require image enhancement techniques for object recognition. Machine learning methods are being increasingly used for object recognition under such adverse conditions. These enhanced object recognition methods of images taken from AUV's has potential applications in underwater pipeline and optical fibre surveillance, ocean bed resource extraction, ocean floor mapping, underwater species exploration, etc. While the classical machine learning methods are very efficient in terms of accuracy, they require large datasets and high computational time for image classification. In the current work, we use quantum-classical hybrid machine learning methods for real-time under-water object recognition on-board an AUV for the first time. We use real-time motion-blurred and low-light images taken from an on-board camera of AUV built in-house and apply existing hybrid machine learning methods for object recognition. Our hybrid methods consist of quantum encoding and flattening of classical images using quantum circuits and sending them to classical neural networks for image classification. The results of hybrid methods carried out using Pennylane based quantum simulators both on GPU and using pre-trained models on an on-board NVIDIA GPU chipset are compared with results from corresponding classical machine learning methods. We observe that the hybrid quantum machine learning methods show an efficiency greater than 65\% and reduction in run-time by one-thirds and require 50\% smaller dataset sizes for training the models compared to classical machine learning methods. We hope that our work opens up further possibilities in quantum enhanced real-time computer vision in autonomous vehicles. 

\end{abstract}

\section{Introduction} \label{sec0}

Underwater imaging is vital for assessing marine ecosystem, evaluating biological and geological habitat {\cite{Chiang2012}}, ocean floor mapping, assessing the health of off-shore oil and natural gas pipelines \cite{Betrand2021,bib1} and under water optical fibres \cite{bib2,bib3,Lu2017,Roberto2020,Jahanbakht2021}. Underwater imaging is mainly achieved by means of specialized cameras and illumination systems built for use in water. The cameras used for underwater imaging are often housed in waterproof housings or enclosures that protect the camera from water damage. Lighting systems are also used to illuminate the subject and improve visibility in the underwater environment. Remotely operated vehicles (ROVs) and autonomous underwater vehicles (AUVs) are used recently for underwater imaging  \cite{Oscar2019, West2001}. ROVs are commanded from the ground and transmitted through direct contact. This restricts its mobility and exposure to isolated regions. Because of their capacity to drive independently using its navigation protocol and environmental knowledge, autonomous underwater vehicles (AUVs) are being used increasingly to capture undersea photographs even from isolated regions. Once deployed, they gather data and return to the ground after the prescribed mission is completed. AUVs are being used for marine biology studies, repair and maintenance, payload delivery to ocean floor, etc, \cite{Robert2018, Marco2015, Simetti2017}. However, capturing high-quality images underwater can be challenging due to light attenuation, water turbulence, and other factors that can affect image clarity and quality\cite{Raihan2019}\cite{Andrey2012}. There are several approaches being used for image enhancement such as histogram equalization (HE) algorithm, Retinex-based methods such as automatic color equalization (ACE) algorithm, Fusion-based methods, etc, \cite{Ahmad2015, Plutino2021, Joshi2008, Ancuti2018}. Physical models include polarization based methods\cite{Schechner2004}, dark-channel prior algorithm \cite{Liu2010} and integral imaging technology \cite{Cho2010} for reconstructing and enhancing the objects in turbid and degraded water. Further, underwater imaging systems typically incorporate stabilization systems, filters, corrective lenses, and machine learning algorithms that can improve image quality and enhance the ability to recognize and analyze underwater objects and environments \cite{Anne2011, Thakur2021}.

\subsection{Machine learning based object-recognition}\label{subsec2}
Machine learning solutions such as object detection and recognition, image enhancement, and underwater sound recognition are being applied recently to improve underwater object detection \cite{Jong2022, Christian2019}.
In this process a large dataset of labelled images containing examples of objects is prepared. Due to the unique characteristics of the underwater environment, such as low visibility, high turbidity, low contrast and high light scattering underwater object recognition becomes extremely challenging \cite{Fan2023}. However, the relevance of underwater robotics and exploration has increased recently, contributing to several developments in underwater object recognition\cite{Frank2020}. 

Algorithms for image improvement typically increase contrast, reduce noise, and boost colour accuracy in underwater photography. Deep learning-based methods such as convolutional neural networks (CNN's), generative adversarial networks (GAN's), etc, improve the quality of underwater images by removing haze and correcting colour distortion \cite{Vilanova2020, Perez2017,Guo2019, Fabri2018,Chen2019}. Integrating machine learning methods can help efficiently navigate and explore underwater surroundings by AUV's. Machine learning techniques can also help in efficient manoeuvering the AUV's by recognising and avoiding obstacles in real-time \cite{Sun2019}, distinguish and categorize underwater noises, such as those made by marine animals, fish, and other species. This can aid researchers in successfully monitoring and studying marine environments\cite{Lowe2022}. However, the main challenge with CNN based methods is that they are computationally expensive and require large datasets for training. Quantum machine learning (QML) based algorithms  are emerging very recently as more efficient compared to their classical counterparts in image classification problems. 

\begin{figure}[ht]
    \centering
    \includegraphics[width=\textwidth]{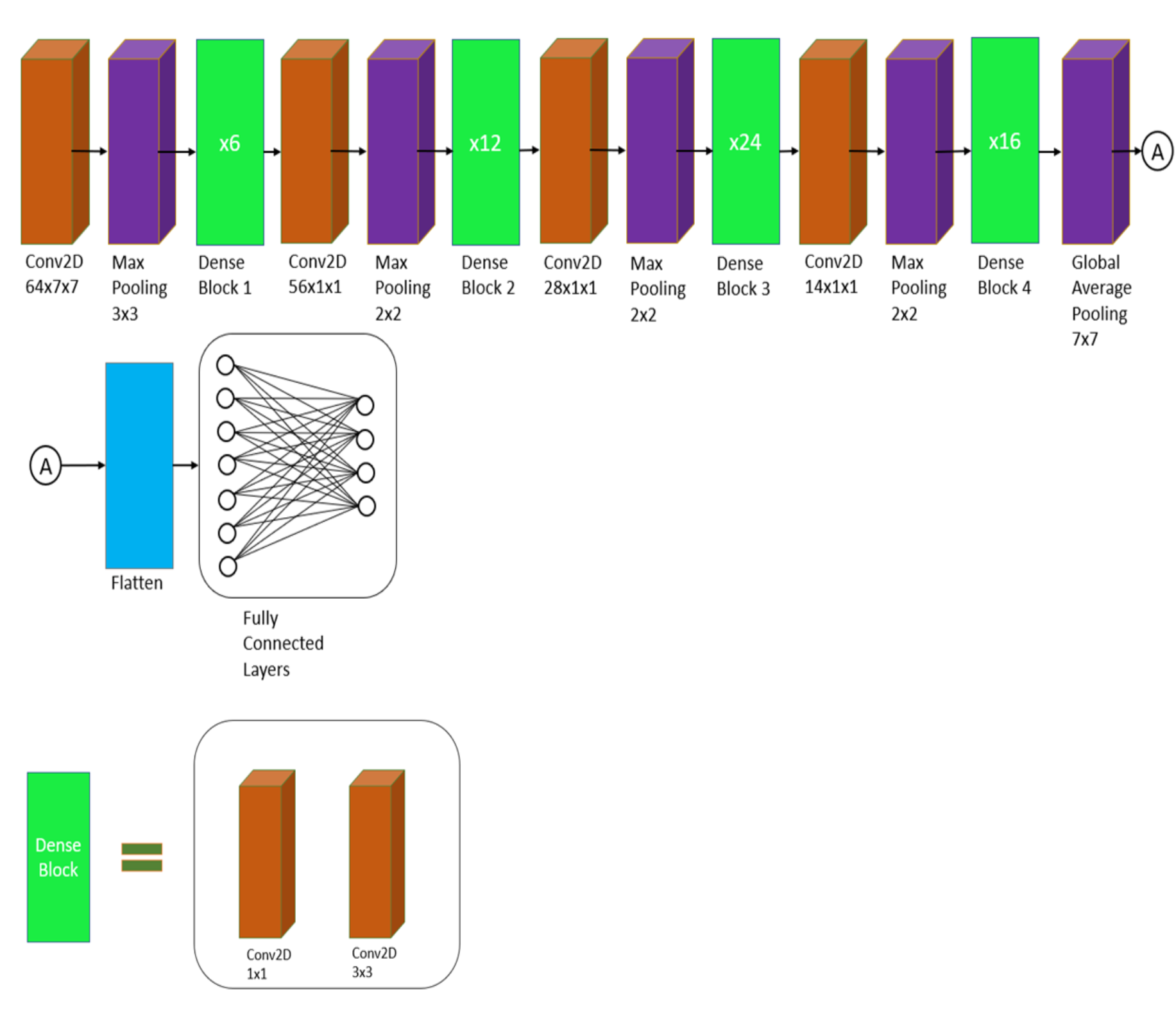}
    \caption{Schematic of Keras DenseNet 121 architecture.}
    \label{Figure 1}
\end{figure}


 In the current article, we apply hybrid classical-quantum machine learning techniques, more specifically, quantum convolutional neural network (QCNN)-based classification methodologies, for real-time and on-surface underwater image classification. We did not find literature on hybrid quantum-based machine learning approaches for real-time image classification of underwater images to the best of our knowledge. This research explores three distinct quantum image encoding mechanisms, namely Quantum Convolutional Neural Network (QCNN)-based on inverse MERA representation, Flexible Representation of Quantum Images (FRQI), and Novel Enhanced Quantum Representation (NEQR), to facilitate the quantum representation of underwater images, respectively \cite{Cong2019, Hirota2011, Zhang2013}. A comprehensive comparative analysis of these methodologies is conducted to evaluate their efficacy and performance comared to the classical CNN methods.
 
\begin{figure}[ht]
    \centering
    \includegraphics[width=\textwidth, height=6cm]{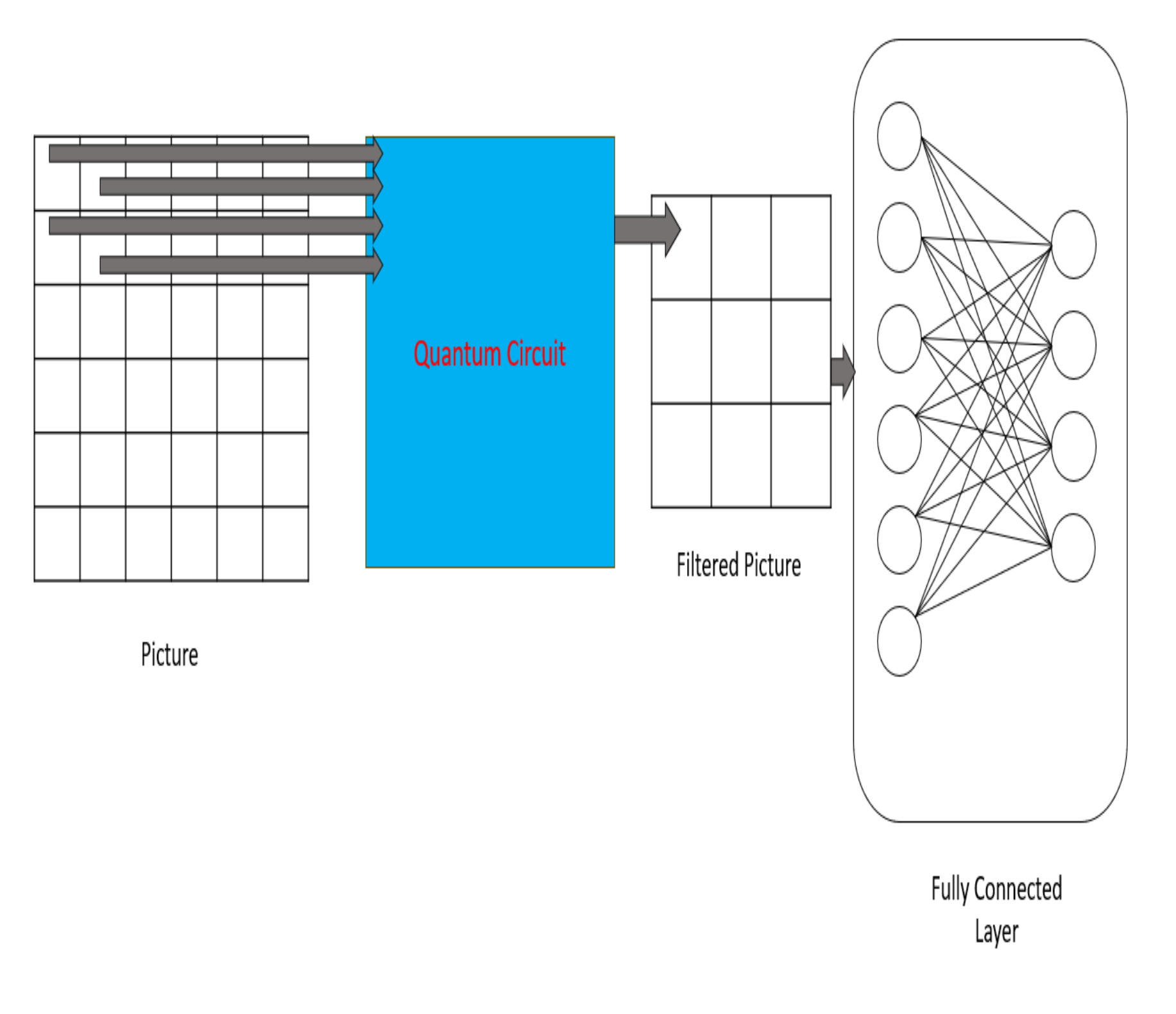}
    \caption{Schematic of hybrid QML architecture.}
    \label{Figure 2}
\end{figure}

 The methods are performed on 5849 images taken from our AUV and 4860 standard Kaggle datasets of sea animals (url in \ref{subsubQS}, \cite{QML_Sea_Animal2023}). The results are compared with the classical CNN-based method using a simple CNN model, standard Keras DenseNet121 and DenseNet201  architectures\cite{huang2017densely}, respectively. The schematic of the DenseNet121 architecture is shown in Figure \ref{Figure 1}. The simple CNN model consists of one convolutional layer \cite{CNN1982}, max-pooling layer \cite{MaxPooling1992} and a ReLU layer \cite{ReLU1969} sequentially, followed by flattening and fully connected layers. Once trained, the algorithm will be deployed on an AUV to autonomously classify images in real-time as the vehicle explores the underwater area. Figure \ref{Figure 2} shows the architecture of hybrid method employed by us with the images encoded using quantum algorithm. The encoded image representations are then measured and sent to fully-connected layers as shown in the figure \ref{Figure 2}. The detailed description of the encoding schemes are given in the following sections.

\subsection{Quantum Convolutional Neural Network method} \label{subsec: QCNN}

\begin{figure}[ht]
    \centering
    \includegraphics[height= 0.3\textwidth, width=0.9\textwidth]{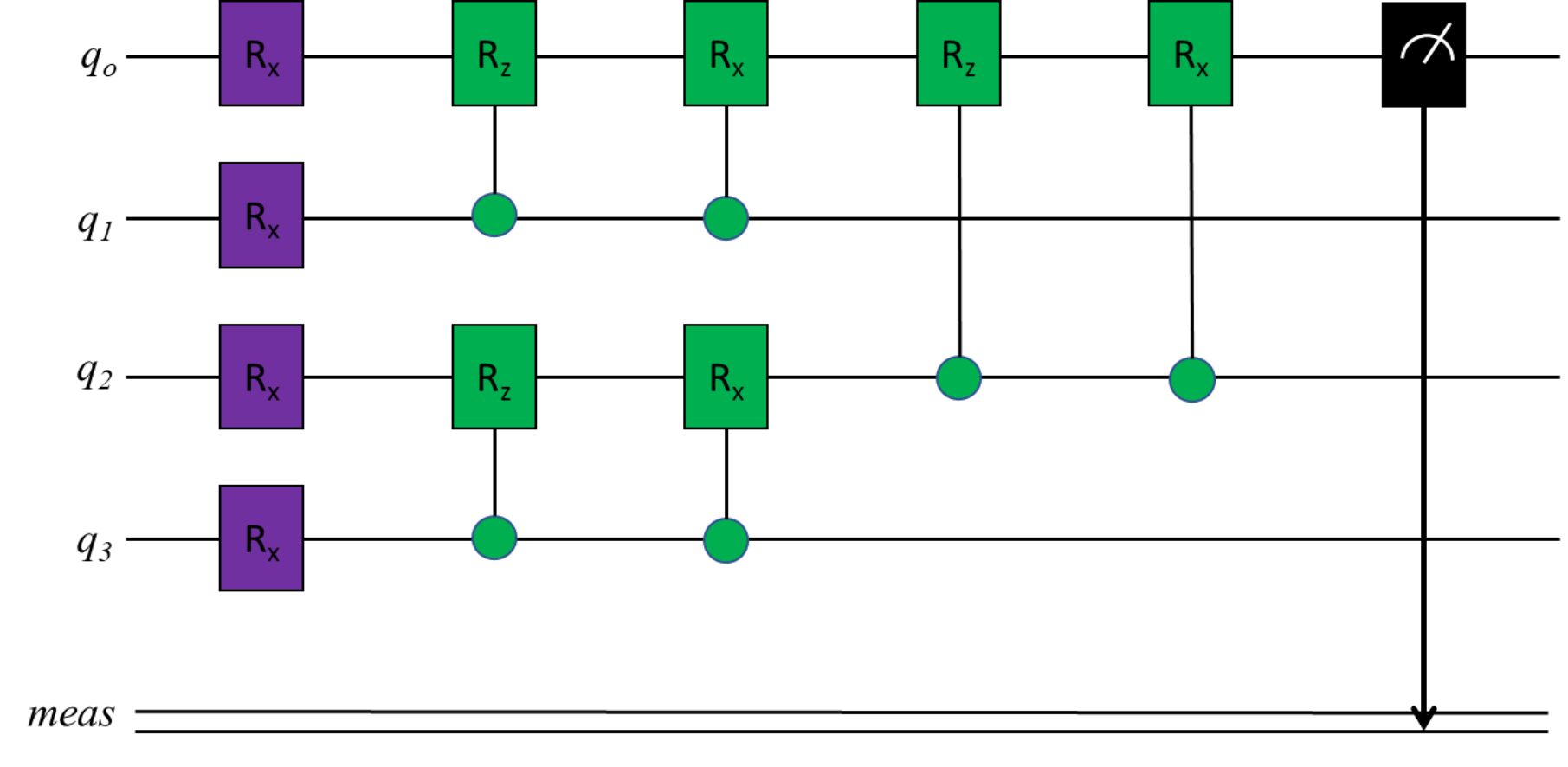}
    \caption{Quantum circuit used in QCNN for the Feature Extraction.}
    \label{Figure 3}
\end{figure}

The input image taken using the on-board camera is sent into a convolutional layer in a QCNN circuit, where a single quasi-local unitary operation ($U_i$) is used \cite{Cong2019, Henderson2020}. The rotations imparted to a set of qubits in pooling layers are determined by measuring neighbouring qubits. The procedure of convolution and pooling continues until the picture sizes are small enough to be fed into fully linked layers. In this paper, we send four pixel values of the input image to four qubits as illustrated in the figure \ref{Figure 3}. Rotation operations $R_x$ followed by permutations of control-$R_z$ gates are applied and measured into a single qubit as shown. Here we employed one convolution and one pooling layer as part of quantum circuit, which are fed into fully connected layers. 
The measurement of qubit is projected onto a specific basis, and the expectation value is calculated as $\langle \psi | \sigma_z | \psi \rangle$, where $\psi$ represents the quantum system's current wave function. 

\subsection{Flexible Representation of Quantum Images} \label{subsec: FRQI}
The flexible representation of quantum images (FRQI) method \cite{Hirota2011} encodes the image in terms of angles $\theta_i$ and pixel positions $\ket{i}$ as shown below.
\begin{equation}
\label{eq1}
\ket{I(\theta)}=\frac{1}{2^n} \sum_{i=0}^{2^{2 n}-1}\left(\cos \theta_i\ket{0} +\sin \theta_i\ket{1}\right) \otimes\ket{i}
\end{equation}

where, $\theta \in\left[0, \frac{\pi}{2}\right]$, $i=0,1,2, \ldots, 2^{2 n}-1 $ and $n$ is number of pixels.
\\
According to the Eq.\eqref{eq1}, the FRQI state is a normalised state with $\lvert\lvert I(\theta) \rvert\rvert=1$ and is composed of two parts:\

\begin{itemize}
    \item color information encoding: $\cos \theta_i\ket{0}+\sin \theta_i\ket{1}$
    \item associated pixel position encoding: $\ket{i}$
\end{itemize}

\begin{figure}[ht]
    \centering
    \includegraphics[height= 0.5\textwidth, width=0.9\textwidth]{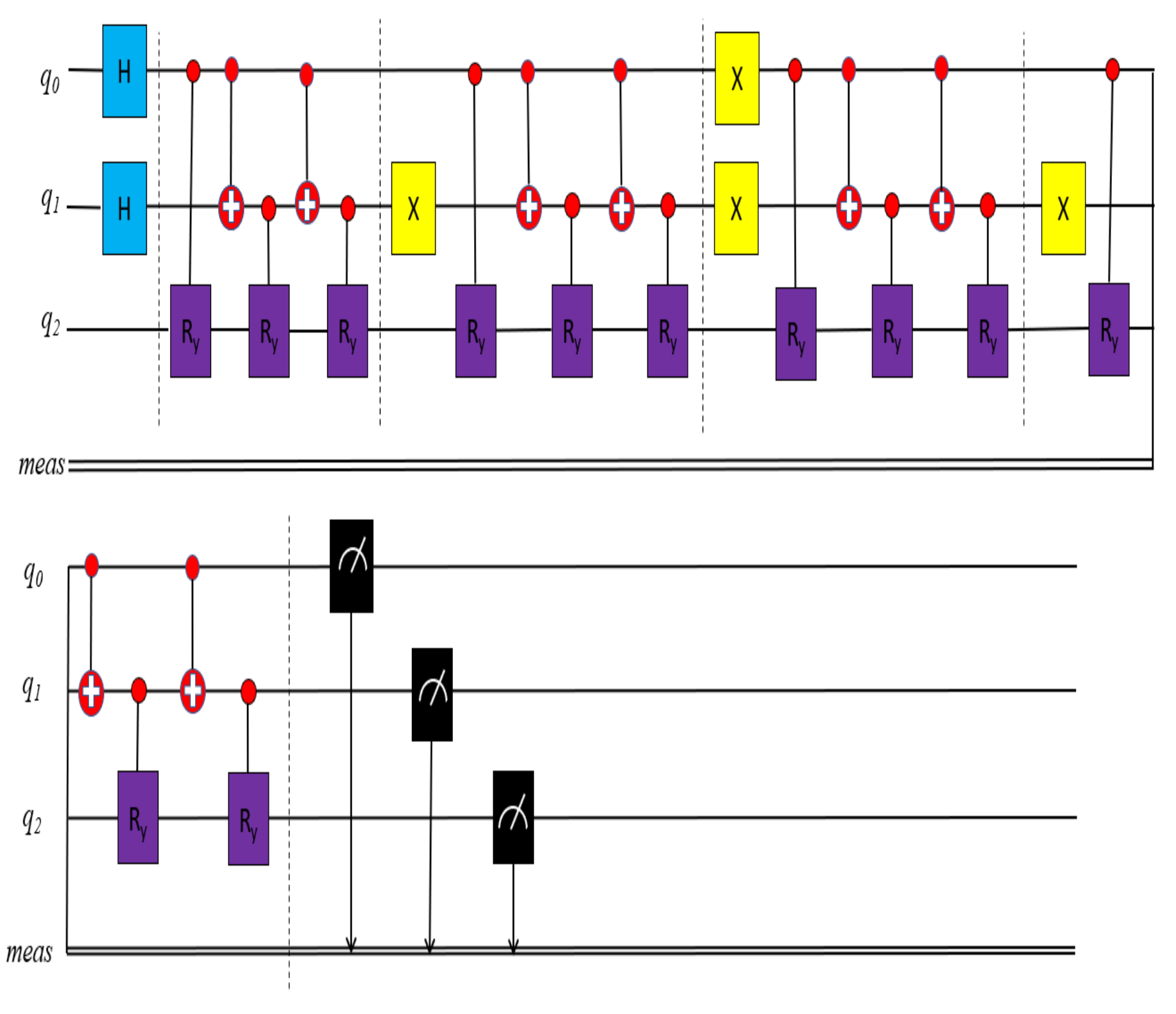}
    \caption{Quantum circuit for FRQI method for eq.\ref{eq2}.}
    \label{Figure 4}
\end{figure}

The quantum circuit using FRQI method is shown in the Fig.\ref{Figure 4}. Unlike QCNN, in FRQI we send single pixel at a time. The color information of the pixel is encoded into $q_2$ in the figure and its position information is sent to $q_0$ and $q_1$, respectively. 

If we consider a $2 \times 2$ pixel image, the intensity ($\ket{I}$) is represented by equation \ref{eq2} and the corresponding circuit is as shown in figure \ref{Figure 4}.

\begin{equation}
\label{eq2}
\begin{aligned}
\ket{I}=\frac{1}{2}[ & \phantom{+} \left(\cos\theta_{0}\ket{0}+\sin\theta_{0}\ket{1} \right)\otimes\ket{00}&\\
& + \left(\cos\theta_{1}\ket{0}+\sin\theta_{1}\ket{1} \right)\otimes\ket{01} \\
& + \left(\cos\theta_{2}\ket{0}+\sin\theta_{2}\ket{1} \right)\otimes\ket{10}\\
& + \left(\cos\theta_{3}\ket{0}+\sin\theta_{3}\ket{1} \right)\otimes\ket{11} \;]
\end{aligned}
\end{equation}

\subsection{Novel Enhanced Quantum Representation} \label{subsec: NEQR}
 FRQI approach, as indicated in the above section \ref{subsec: FRQI} uses a single qubit to retain gray-scale values and spatial image location, which is inefficient for image encoding and extraction. Novel Enhanced Quantum Representation (NEQR) \cite{Zhang2013} is an improved variant of FRQI wherein the pixel color is represented in 8 qubits with values ranging from 0 to 255 and the pixel locations in 2 qubits as illustrated in the figure \ref{Figure 5}. $P_1$ and $P_2$ in the figure \ref{Figure 5} represent position of the pixels and the qubits $I_1$ to $I_7$ represent gray-scale values of each pixel. The intensity values of an image represented by quantum bits $|I\rangle$ is given by the following equation:
 
\begin{figure}[ht]
    \centering
    \includegraphics[height= 0.6\textwidth, width=0.9\textwidth]{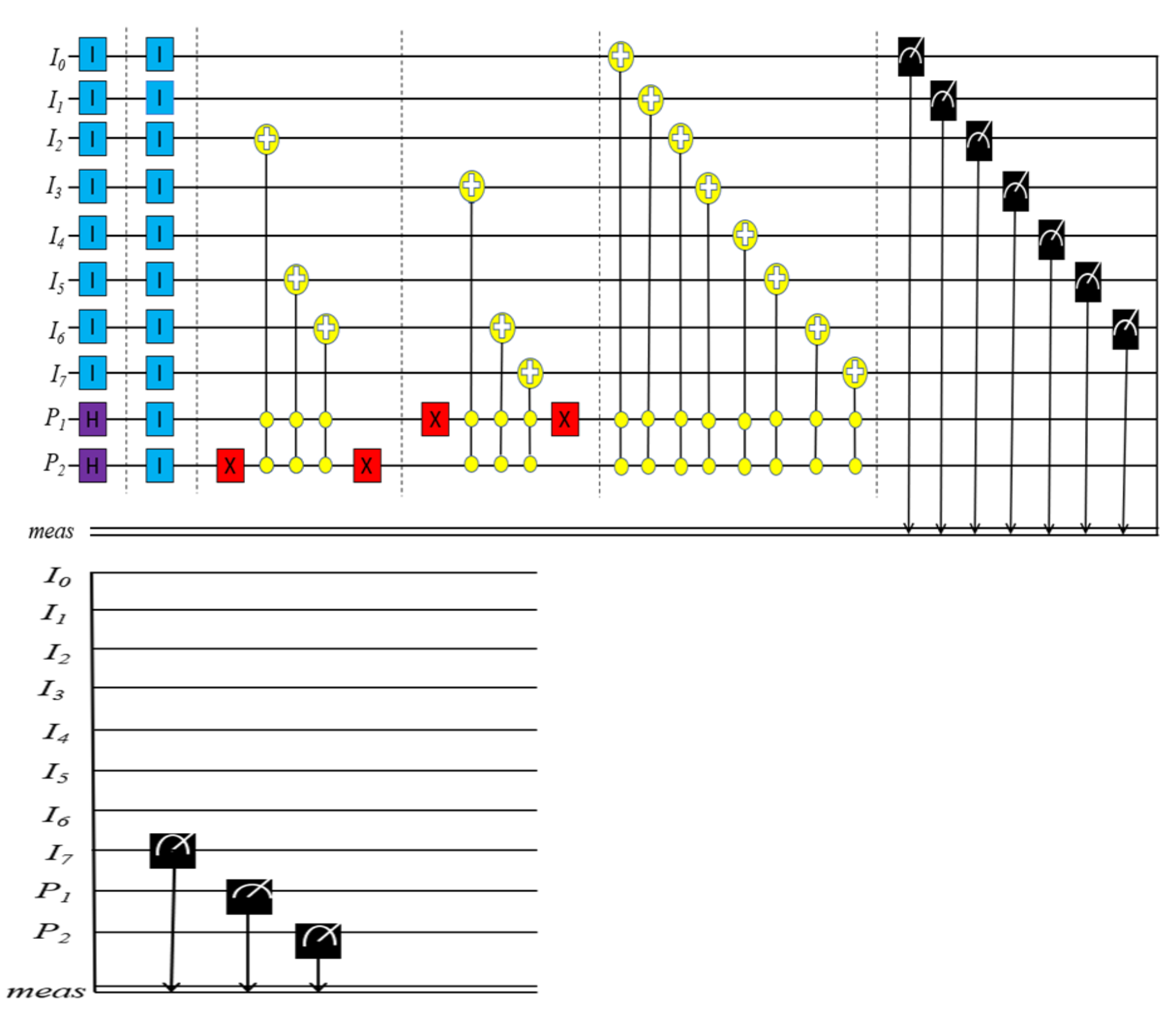}
    \caption{Quantum circuit for NEQR method for eq.\ref{eq3}.}
    \label{Figure 5}
\end{figure}

\begin{equation}
\label{eq3}
\ket{I} = \frac{1}{2^{n}}\sum_{Y=0}^{2^{2n-1}} \sum_{X=0}^{2^{2n-1}} \ket{f(Y,X)}\ket{YX} 
\end{equation}
where $\ket{f(Y,X)}$, $\ket{YX}$ and $n$ represent pixel intensity value, pixel position co-ordinates and number of pixels, respectively. 
Gray-scale images of size $512 \times 512$ are considered in our work.



\section{Results and Discussions}\label{sec2}

The Quantum encoded images using methods described in the earlier section are then loaded into fully connected layers. The codes are implemented using Jupyter notebooks and neural network models are executed in TensorFlow 2 and Keras. Xanadu Technology's PennyLane simulator is used to accommodate the quantum layers.
A random number generator is utilised to initialise the weights and biases, and normalisation tools such as dropout and embedding procedures are employed to determine the foundation for stochastic gradient. 
Training of the model is performed using a training dataset of 4679  images taken in murky water, clear water, enhanced images using GAMMA correction method \cite{Guan2009} and 3888 standard Kaggle images in each round of our inquiry. 

Training and validation losses are calculated at the conclusion of each epoch. The estimated accuracy and loss are calculated as shown below.

\begin{equation}
    \label{eq5}
    Accuracy = \frac{True_{positive} + True_{negative}}{True_{positive} + True_{negative} + False_{positive} + False_{negative}}
\end{equation}

 \begin{equation}
    \label{eq6}
    Loss = -\sum_{i=1}^{output size} y_{i}\log{\hat{y_{i}}}
\end{equation}

\subsection{On-Surface object detection}\label{subsec6}

\subsubsection{Quantum Simulator} \label{subsubQS}
\begin{figure}[ht]
    \centering
    \includegraphics[height=0.55\textwidth,width=0.9\textwidth]{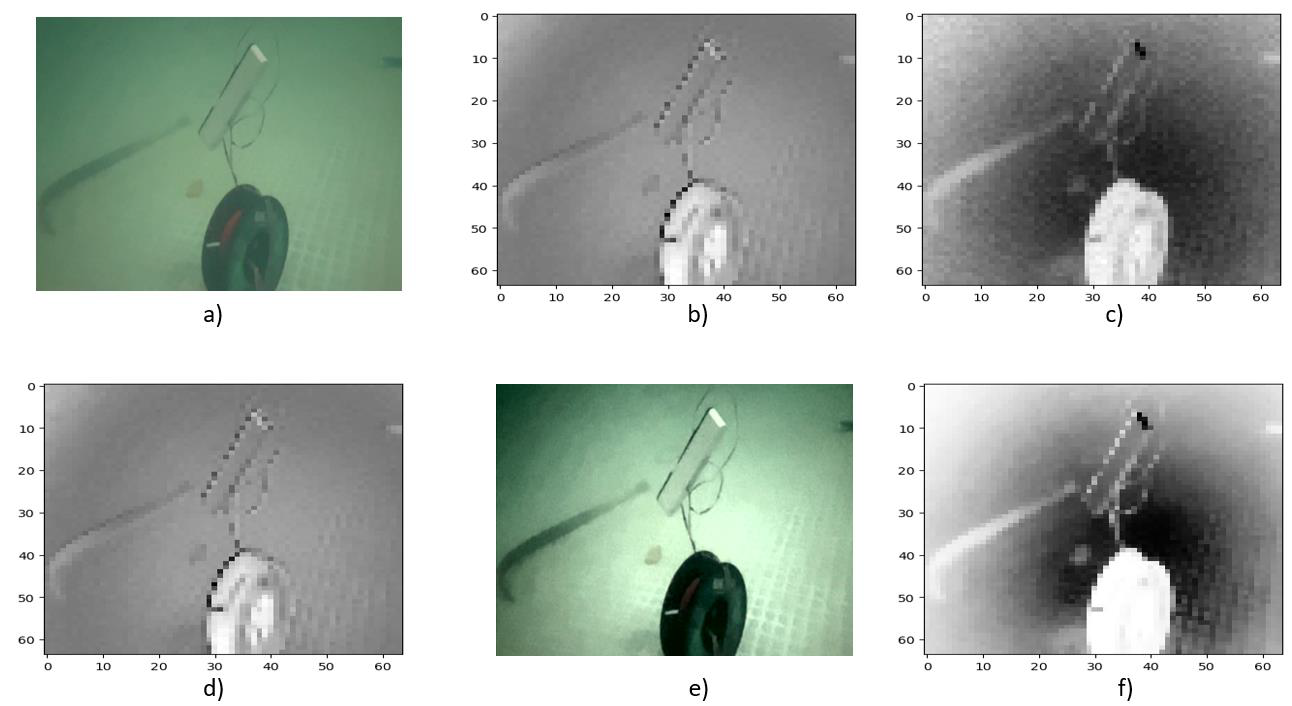}
    \caption{(a) Images of a Submerged object taken from AUV in murky water and its image representation based on (b) Inverse MERA [QCNN], (c) FRQI, (d) NEQR, (e) Contrast Enhanced Image, and (f) QCNN of Contrast Enhanced Image (respectively).}
    \label{Figure 6}
\end{figure}

\begin{figure}[ht]
    \centering
    \includegraphics[height=0.5\textwidth,width=0.8\textwidth]{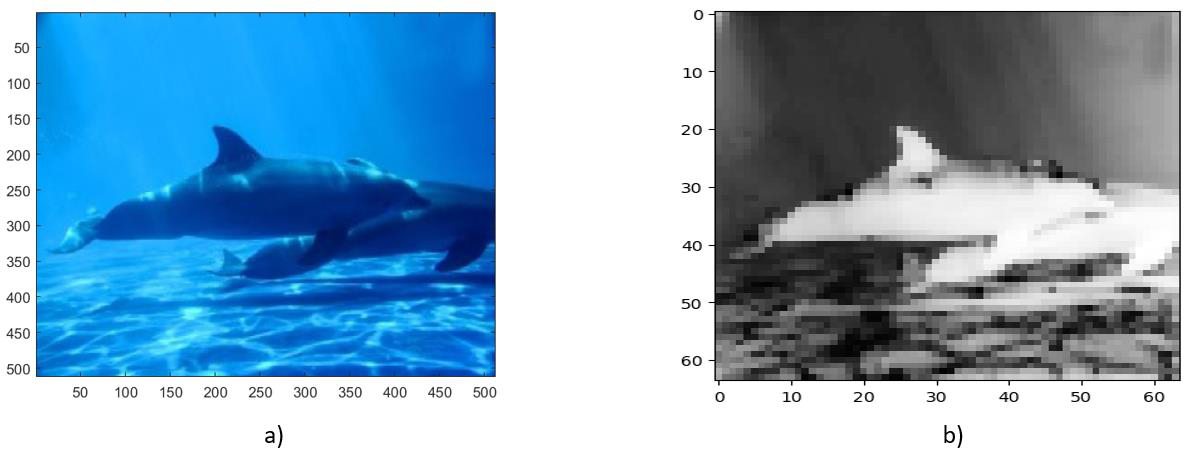}
    \caption{(a) Image of Dolphin taken from KAGGLE and its (b) QCNN image representation.}
    \label{Figure 7}
\end{figure}

The images captured by AUV are taken to the surface where the script is implemented on a NVIDIA DGX-1 server with 1 TB RAM, 15TB SSD, 128 CPU cores, and 1 $\times$ A100(Volta) GPUs using Ubuntu 20.04 OS. Figure \ref{Figure 6} represents an image taken from AUV, its enhanced version and their quantum representations. The contrast enhancement of the images taken from AUV is performed by using GAMMA correction algorithm \cite{Guan2009} in MATLAB, as shown in Fig. \ref{Figure 6}e. Gamma correction is a nonlinear operation used in multimedia or static photographic systems to encode and decode luminance values. The value of $\gamma$ taken to increase the contrast is 0.5. We further performed the image classification of images taken from a standard dataset, i.e., sea animals from \url{https://www.kaggle.com/datasets/vencerlanz09/sea-animals-image-dataste} . We used only 10 classes of sea animals from Kaggle dataset. The dataset taken from AUV are also labelled into 10 different classes. Figure \ref{Figure 7}a shows the original image taken from kaggle dataset and its  QCNN image representation in Figure \ref{Figure 7}b.

\begin{figure}[ht]
    \centering
    \includegraphics[height=0.8\textwidth,width=0.9\textwidth]{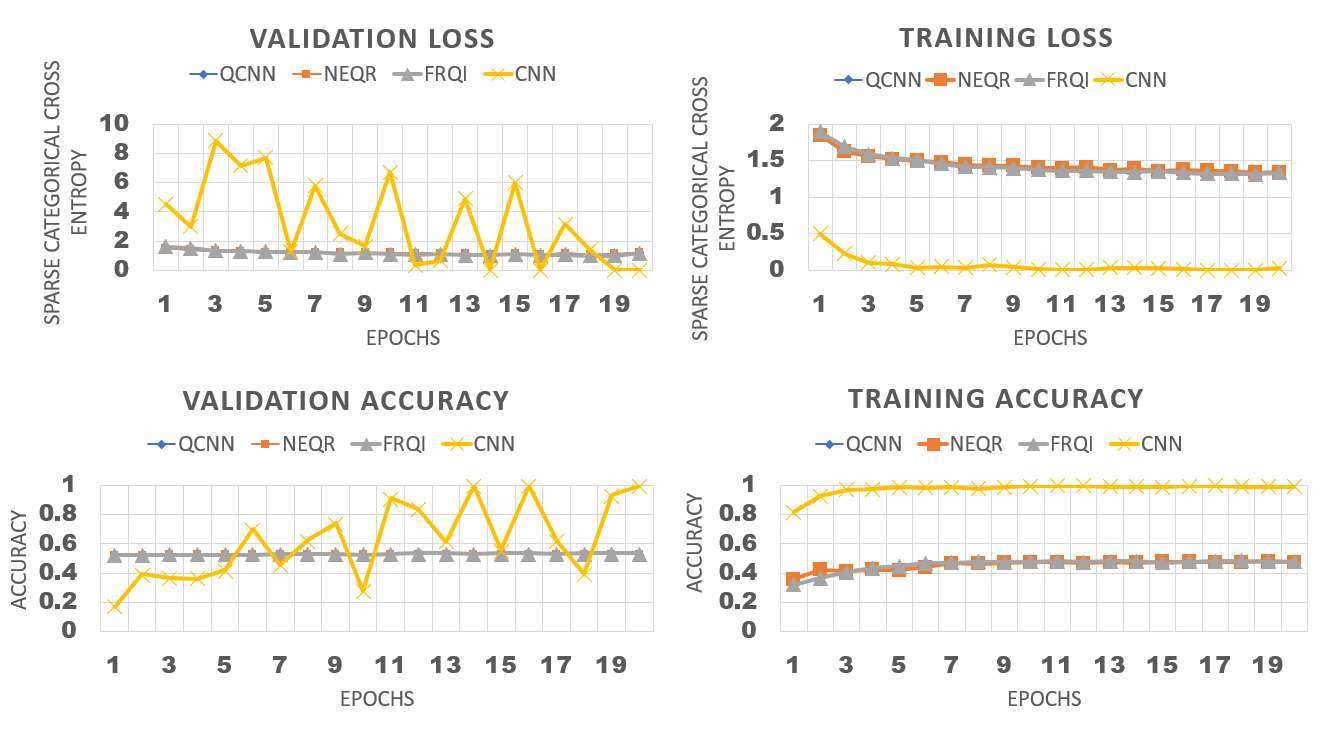}
    \caption{Comparative analysis of training and validation losses for classical CNN, Inverse MERA (QCNN) FRQI and NEQR respectively of low light underwater images taken in murky water from Quantum Simulator.}
    \label{Figure 8}
\end{figure}

\begin{figure}[ht]
    \centering
    \includegraphics[height=0.8\textwidth,width=0.9\textwidth]{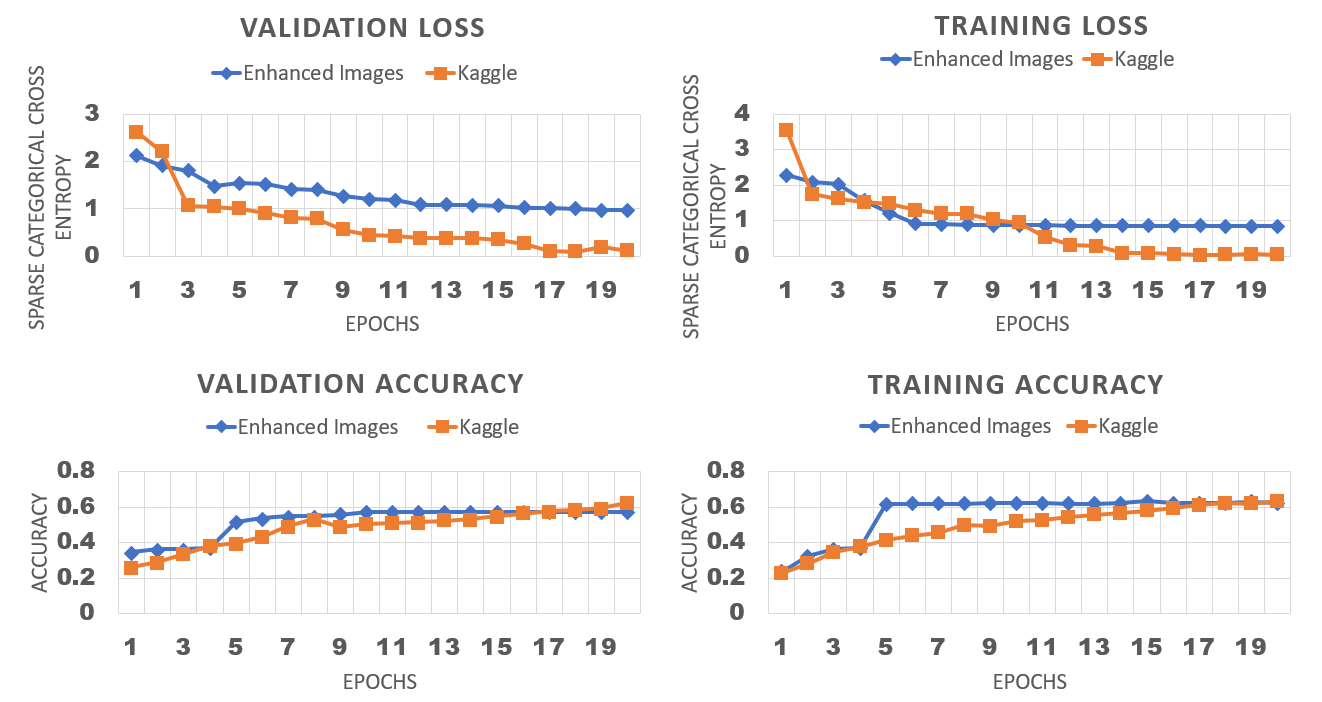}
    \caption{Comparative analysis of training and validation losses for QCNN of contrast enhanced and Kaggle images in Quantum Simulator.}
    \label{Figure 9}
\end{figure}


Figures \ref{Figure 8} and \ref{Figure 9} show loss and accuracy of sparse categorical cross entropy versus the number of epochs for low-light images (fig \ref{Figure 8}), enhanced images (fig \ref{Figure 9}) and image datasets from Kaggle (fig \ref{Figure 9}), respectively. From all the above figures it is observed that the validation and training losses are very less and quickly saturate in case of hybrid methods compared to the classical CNN method, i.e., DenseNet121. Validation and training accuracies typically saturate after 4 epoch cycles in the case of hybrid algorithms in contrast to DenseNet121 as seen from the figures \ref{Figure 8} and \ref{Figure 9}.

Table \ref{tab1} shows the prediction accuracy, loss and runtime for classical CNN based methods - simple CNN, DenseNet121, DenseNet201 and hybrid quantum-classical methods- QCNN, FRQI and NEQR methods for the underwater images taken in murky waters and clear water, respectively. Runtime here are taken as time taken for both training and validation.
It is observed that while classical CNN methods achieve more accuracy and less losses, hybrid methods clearly achieve much shorter run-times. The accuracies achieved are slightly higher ($70\%$) for the clear waters compared to the murky waters in the case of QCNN method. 
We further performed the classification using 2046 QCNN encoded images with $70\%$ used for training data and $30\%$ used for prediction and observed that the accuracies and losses has not changed proving that hybrid quantum image classification requires lesser datasets than classical CNN methods.

The results shown are simulated using PennyLane simulator and may differ when used on real quantum hardware. The error correction schemes are not implemented for any of the quantum image representation methods. It is reported that QCNN method naturally incorporates the quantum error correction in the convolution layers \cite{Cong2019}. The models trained in NVIDIA DGX-1 on-surface are used as pre-trained models for real-time onboard object detection in AUV shown in Subsection \ref{subsec7}.

\begin{table}[htbp]
\centering
\caption{\label{tab1}Training performance comparison between classical and hybrid models for murky and clear water image dataset (clear water indicated in brackets).}
\footnotesize 
\begin{tabular}{>{\raggedright\arraybackslash}p{0.18\textwidth} 
                >{\centering\arraybackslash}p{0.1\textwidth} 
                >{\centering\arraybackslash}p{0.1\textwidth} 
                >{\centering\arraybackslash}p{0.15\textwidth} 
                >{\centering\arraybackslash}p{0.15\textwidth}}
\toprule
\textbf{Models} & \multicolumn{4}{c}{\textbf{Metrics}} \\
\cmidrule{2-5}
& \textbf{Images} & \textbf{Acc.} & \textbf{Loss} & \textbf{Runtime} \\
\midrule
CNN (murky water images) & 4094 & 94\% & 0.021 & 10:26 min \\
DenseNet121 (murky water images) & 4094 & 90\% & 0.1544 & 19:10 min \\
DenseNet201 (murky water images) & 4094 & 80\% & 0.249 & 27:18 min \\
QCNN (murky water images) & 4094 & 60\% & 0.9956 & 3:26 min \\
FRQI (murky water images) & 4094 & 52\% & 1.2 & 4:05 min \\
NEQR (murky water images) & 4094 & 59\% & 1.156 & 9:43 min \\
DenseNet121 (clear water images) & 3462 & 96\% & 0.0645 & 18:43 min \\
QCNN (clear water images) & 3462 & 71\% & 0.6485 & 2:54 min \\
\bottomrule
\end{tabular}
\end{table}



\subsubsection{Quantum Hardware} \label{subsubQH}
We perform the image encoding using quantum representations on real quantum hardware which are then sent to fully connected layers.
We implement this hybrid method by taking the images from AUV to the surface where the quantum circuit is run on a real NISQ devices to report its performance for accurately detecting the objects. We employed QCNN algorithm with 4 qubits on \textit{ibm\_brisbane}. The \textit{ibm\_brisbane} is 127-qubit quantum processor based on superconducting circuits. We used Pennylane-Qiskit plugin which helps to integrate our circuit written in Pennylane to run on IBM Quantum Hardware. The code has been transpiled to run on \textit{ibm\_brisbane}. The optimization level used on the hardware is 1 with dynamic error suppression and 1Q gate optimization with 10000 shots. As with quantum simulator there is saturation in losses and accuracy even in real quantum hardware after 4 epoch cycles. The choice of QCNN method as shown in Table \ref{tab1} is because of its higher accuracy and lower runtime than NEQR and FRQI methods.

\begin{figure}[htbp]
    \centering
    \includegraphics[height=0.4\textwidth,width=0.9\textwidth]{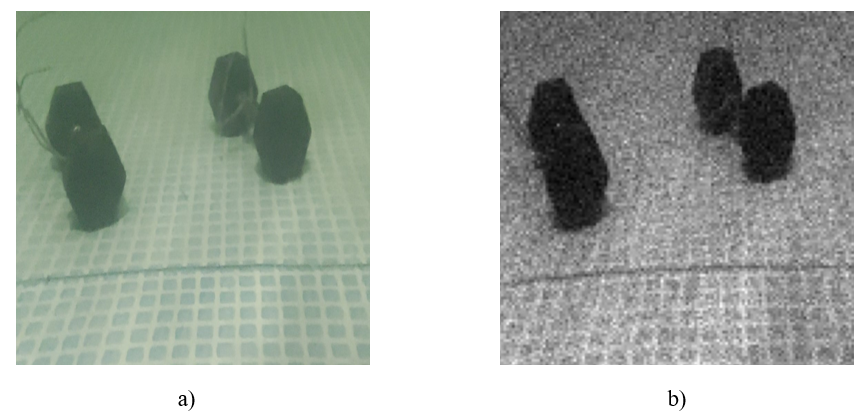}
    \caption{Image of a Submerged object taken from AUV in clear water and its image representation based on (b) Inverse MERA [QCNN] from IBM Quantum Hardware.}
    \label{fig:10}
\end{figure}

\begin{figure}
    \centering
    \includegraphics[height=0.4\textwidth,width=0.9\textwidth]{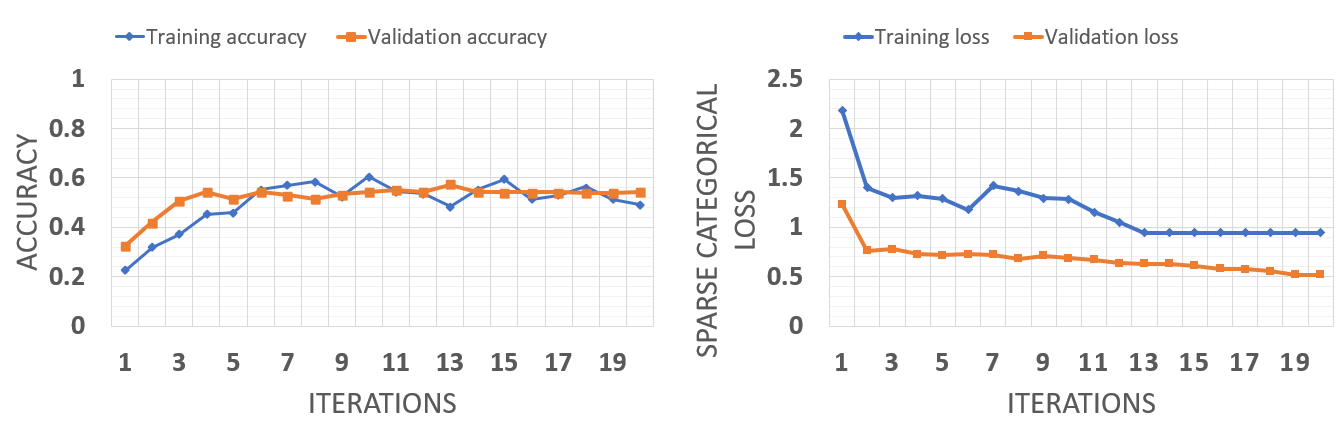}
    \caption{Analysis of training and validation loss and accuracy of images taken in clear water for QCNN method in IBM Quantum Hardware.}
    \label{fig:11}
\end{figure}

From figure \ref{fig:10} we can see that the QCNN image representation through quantum hardware is not as clear as in quantum simulator. The image quality is effected by the noise introduced in the quantum hardware. Due to this the accuracy in training and validating the model is decreased by about 13\% relative to quantum simulator as shown in figure \ref{fig:11}. When the same dataset is sent through a quantum simulator we can validate the model with an accuracy of around 68\%. Table \ref{tab2} show the loss and accuracy obtained from the IBM Quantum Hardware with the QCNN approach for the clear and murky water images. The runtime per image for quantum encoding is 10 seconds for the quantum computer. 

\begin{table}[htbp]
\caption{\label{tab2}Performance of hybrid QCNN model on quantum hardware with images in clear water and murky water.} 
\centering
\begin{tabular}{lcc}
\toprule
\multirow{2}{*}{\textbf{Images}} & \multicolumn{2}{c}{\textbf{Metrics}} \\
\cline{2-3}
 & \textbf{Loss} & \textbf{Accuracy} \\
\midrule
Clear Water Images & 0.5143 & 56.21\% \\
Murky Water Images & 0.993 & 46.2\% \\
\bottomrule
\end{tabular}
\end{table}

\subsection{Real-time On-board object detection} \label{subsec7}
\begin{figure}[ht]
    \centering
    \includegraphics[width=1\textwidth]{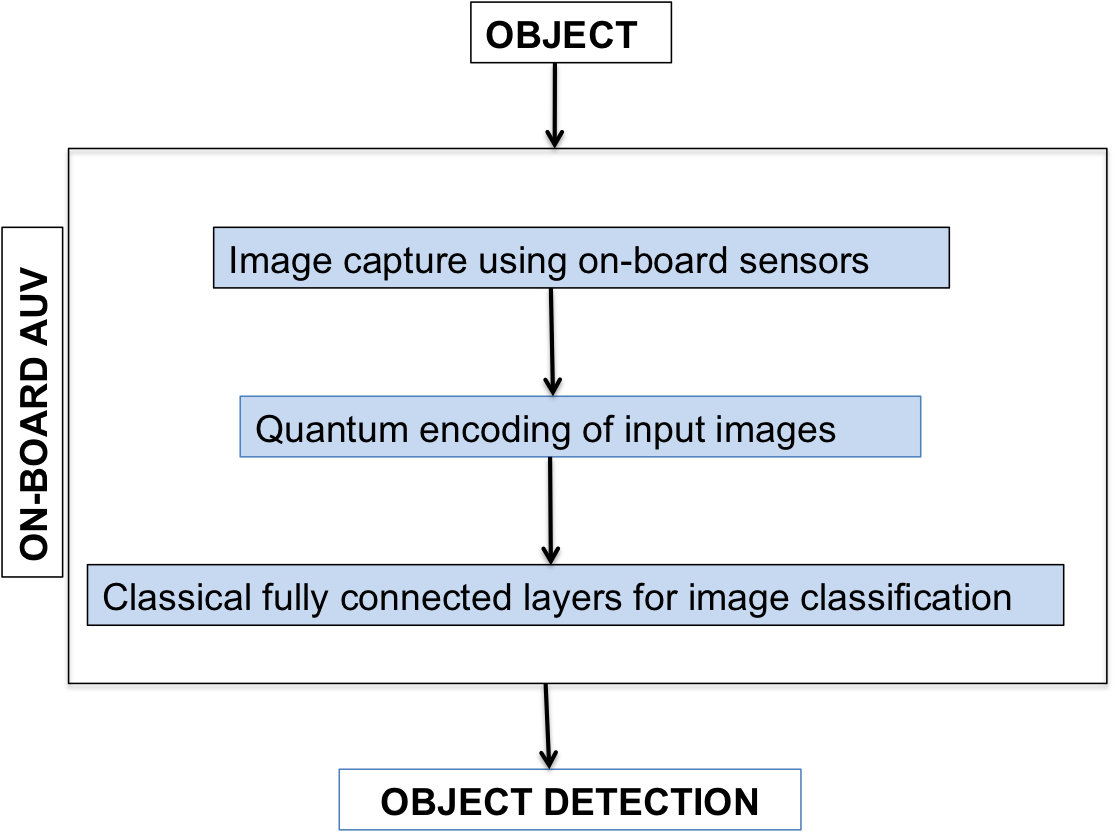}
    \caption{Flowchart of the on-board object detection using hybrid quantum-classical CNN algorithm.}
    \label{Figure 12}
\end{figure}

Further, we perform the hybrid quantum-classical CNN method right from image acquisition to classification on-board an AUV as illustrated in the figure \ref{Figure 12}. An AUV is built from scratch with an image sensor integrated NVIDIA Jetson Xavier NX module 8 GB RAM, 512 GB SSD, 6 ARMv8 CPU processors, 384 core Volta architecture using Ubuntu 20.04 on board with AUV. The CPU of the Jetson Xavier module processes the images taken by camera on-board AUV. 600 images were taken and sent to the on-board NVIDIA Jetson CPU for encoding and predicting the objects real-time. In the real-time object detection, we used pre-trained model of both QCNN and DenseNet121 in clear water and murky water as stated in Section \ref{subsubQS} for prediction.

\begin{figure}
    \centering
    \includegraphics[width=0.9\textwidth]{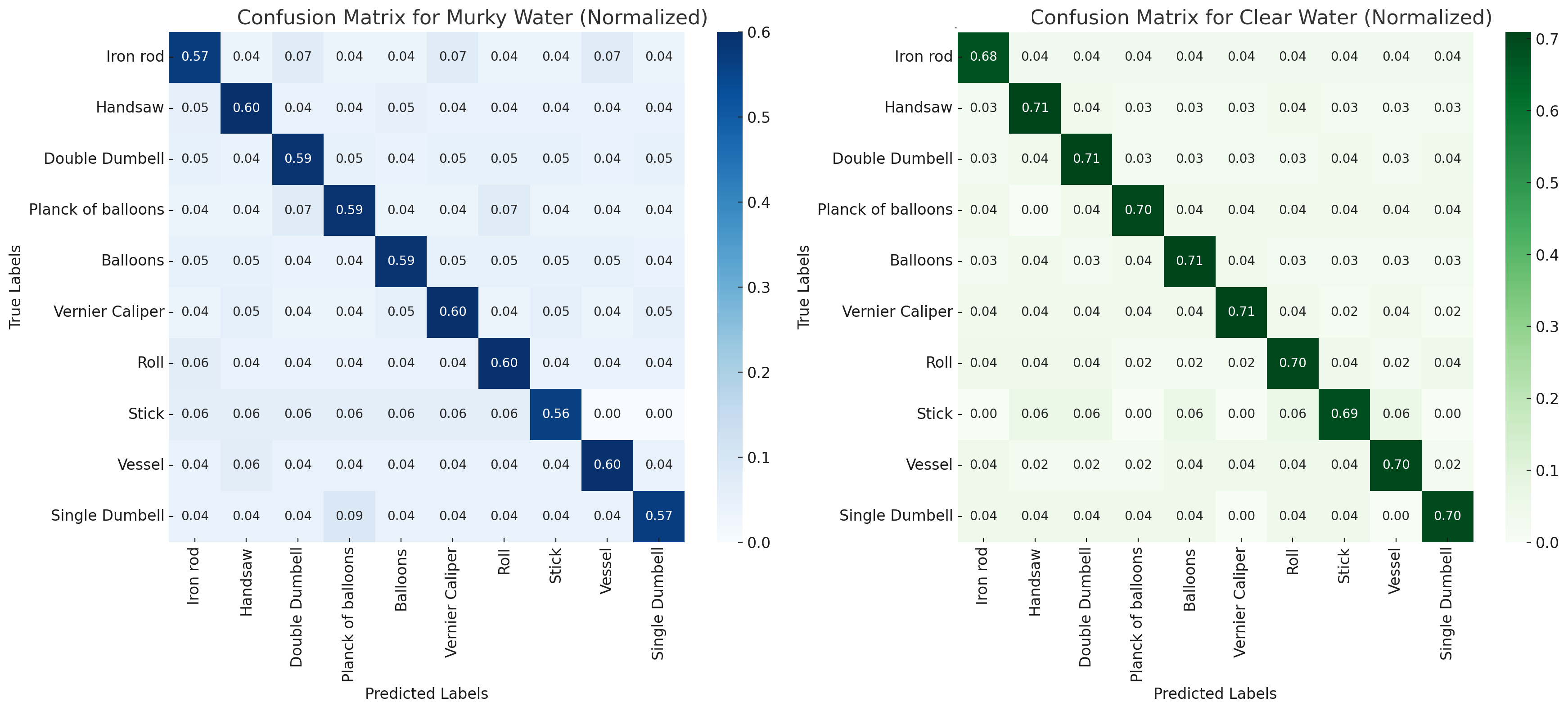}
    \includegraphics[width=0.9\textwidth]{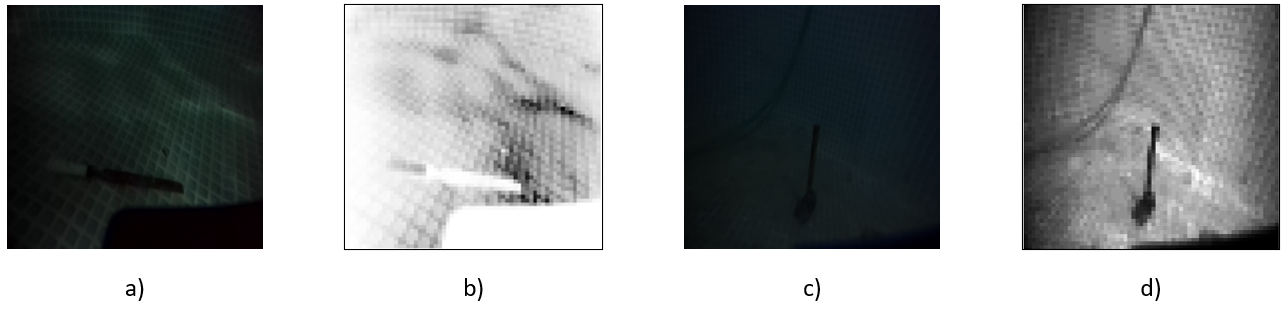}
    \caption{Images of objects taken in (a) low-light clear water and (b) its QCNN representation, (c) murky water and(d) its QCNN representation performed on-board moving AUV along with the confusion matrix for QCNN models.}
    \label{fig:13}
\end{figure}

\begin{table}[ht]
\caption{\label{tab3}Performance of the QCNN model and DenseNet121 model onboard AUV with real-time images in clear and murky water. (Training done in CPU mode)}
\centering 
\begin{tabular}{lcccc}
\toprule
\multirow{2}{*}{\textbf{Metrics}} & \multicolumn{2}{c}{\textbf{QCNN}} & \multicolumn{2}{c}{\textbf{DenseNet121}} \\
\cmidrule(lr){2-3} \cmidrule(lr){4-5}
& \textbf{Clear Water} & \textbf{Murky Water} & \textbf{Clear Water} & \textbf{Murky Water} \\
\midrule
Accuracy & 60\% & 52\% & 81\% & 73\% \\
Encoding Time & 19 sec & 20 sec & --- & --- \\
Runtime & 200 msec  & 200 msec & 2 sec & 2 sec \\
\bottomrule
\end{tabular}
\end{table}

The runtimes mentioned for QCNN and DenseNet121 are the times taken to predict per image using on-board GPU on our AUV. In the QCNN model, the majority of the computational time is attributable to the process of encoding the image using a quantum circuit, which is the most time-intensive step as shown in Table \ref{tab3}. For images in clear water, the accuracy of the pre-trained model onboard the AUV is 60\% while in murky water it is 52\%. The confusion matrices depicted in figure \ref{fig:13} show the performance of the hybrid algorithm in clear and murky waters. The decrease in accuracies for predicting images using onboard classification can be attributed to additional motion-blur from movement of AUV. Table \ref{tab3} shows the performance of QCNN method as compared to the DenseNet121 method for on-board object detection. While QCNN method shows 35\% and 40\% lesser accuracies than the DenseNet121 method for clear and murky water, respectively, the run times are 460\% and 510\% shorter than the corresponding classical method as seen from the table 3. We have considered DenseNet121 method for the comparison in the table3, however the other methods mentioned earlier also showed similar results in accuracies and run-times. 

\section{Conclusion}\label{sec13}

In this article we report detailed classification of underwater images using hybrid quantum CNN method. We take the route of performing the quantum encoding of classical images and the measured values are sent to classical fully connected layers. We compare three existing quantum image encoding schemes QCNN (inverse MERA representation), FRQI and NEQR methods. QCNN method shows higher accuracy and faster run times amongst the methods used for the low-light, low-contrast images taken. The classification is performed on GPU servers and compared with classical CNN methods. In all the cases classical CNN methods performed better while the run times of hybrid quantum CNN methods are significantly higher. We observed that the QCNN encoded images require lesser datasets for classification as opposed to classical CNN methods. We further deployed the pre-trained hybrid QCNN method on-board AUV and implemented real-time object classification. The hybrid QCNN methods showed accuracies of about 60\% and the image classification run-time per image, which is taken as the total time right from the image capture till the classification per image is within 19 seconds. The run-times are about 3 times faster than the DenseNet121 method on-board AUV. We also compared performances of various other classical CNN's with the hybrid methods and observed in all the cases hybrid methods out-performed the classical methods in computational times while the classical methods were more accurate in detection. However the on-board simulations were implemented on the CPU's of Xavier chip and run-times can be further improved if we could use the GPU's which was not possible due to dependency issues with the Pennylane simulator. The accuracies can be further improved by training the models with enhanced or clear images as well as by data augmentation using GAMMA correction methods, sonar signals, etc.  
Hence we propose that hybrid QML algorithms, particularly QCNN method, can be used for real-time object detection and manoeuvre of any unmanned vehicles like unmanned aerial vehicles (UAVs), AUVs, autonomous vehicles, etc and we believe that our work paves way for real-time use cases of hybrid quantum algorithms in computer vision in the current NISQ era.

\ack
This study is based on research that has been supported and financed by Mahindra University.

\section*{References}

\bibliography{JPCSLaTexGuidelines}

\end{document}